\documentclass[letterpaper, 10 pt, conference]{ieeeconf}

\IEEEoverridecommandlockouts                              

\usepackage[l2tabu,orthodox]{nag}

\usepackage[
backend=biber, 
bibencoding=utf8,
style=ieee, 
sorting=none, 
natbib=true, 
doi=false, 
isbn=false, 
url=false, 
eprint=false, 
maxcitenames=1, 
mincitenames=1,
]{biblatex}

\usepackage[draft,pdftex,colorlinks]{hyperref}

\usepackage[printonlyused]{acronym}

\usepackage{siunitx}
\usepackage[all]{nowidow}

\usepackage[dvipsnames]{xcolor}

\usepackage{lipsum}

\usepackage[pdftex]{graphicx}

\usepackage{epstopdf}

\usepackage{import}

\graphicspath{{./latexGoodPractices/}}

\usepackage{booktabs}

\usepackage{tabu}

\usepackage{amssymb,amsfonts,amsmath,amscd}

\usepackage{bm} 

\newcommand{\norm}[1]{\left\Vert#1\right\Vert}

\newcommand{\bbm}{\begin{bmatrix}}
\newcommand{\ebm}{\end{bmatrix}}

\usepackage[normalem]{ulem} 

\makeatletter
\usepackage[utf8]{inputenc}
\usepackage{commath}
\usepackage{mathtools}
\usepackage{mathabx}
\usepackage{siunitx}
\usepackage[ruled,vlined]{algorithm2e}
\usepackage[lofdepth,lotdepth]{subfig}
\usepackage{tikz}

\let\oldtheequation\theequation
\renewcommand\tagform@[1]{\maketag@@@{\ignorespaces#1\unskip\@@italiccorr}}
\renewcommand\theequation{(\oldtheequation)}
\makeatother

\usepackage[belowskip=0pt,aboveskip=1pt]{caption}  
\graphicspath{{media/}}

\newcommand{\ie}{i.e., }

\newcommand{\SE}{\mathrm{\mathbf{SE}}}

\newcommand{\se}{\mathfrak{se}}

\newcommand{\bphi}{\bm{\phi}}

\newcommand{\bxi}{\bm{\xi}}

\newcommand{\R}{\mathbb{R}}

\DeclareMathOperator*{\argmin}{arg\,min}

\usepackage{amssymb}
\usepackage{pifont}
\newcommand{\vzero}{\bm{0}}

\acrodef{lidar}{Light Detection And Ranging}
\acrodefplural{lidar}{Light Detection And Ranging}
\acrodef{ICP}{Iterative Closest Point}
\acrodef{DOF}{Degrees Of Freedom}
\acrodef{GP}{Gaussian Process}
\acrodefplural{GP}{Gaussian Processes}
\acrodef{RTS}{Robotic Total Station}
\acrodefplural{RTS}{Robotic Total Stations}
\acrodef{DOF}{Degrees Of Freedom}
\acrodef{GNSS}{Global Navigation Satellite System}
\acrodef{RTK}{Real Time Kinematics}
\acrodef{SVD}{Singular Value Decomposition}
\acrodef{RP}{Reference Point}
\acrodef{GCP}{Ground Control Point}
\acrodef{UAV}{Unmanned Aerial Vehicle}
\acrodef{IQR}{Interquartile Range}

\DeclareUnicodeCharacter{0308}{¨} 

\begin{document}
\title{\LARGE \textbf{Extrinsic calibration for highly accurate trajectories reconstruction}}

\author{Maxime Vaidis$^{1}$, William Dubois$^{1}$, Alexandre Guénette$^{1}$, Johann Laconte$^{1}$, \\Vladim\'ir Kubelka$^{2}$, François Pomerleau$^{1}$
\thanks{$^{1}$Northern Robotics Laboratory, Université Laval, Québec City, Canada,
  {\texttt{\small{$\{$maxime.vaidis, francois.pomerleau$\}$ @norlab.ulaval.ca}}}}%
\thanks{$^{2}$Mobile Robotics and Olfaction lab of the AASS research center at Örebro University, Sweden.}%
}

\linepenalty=3000
\addtolength{\textfloatsep}{-0.1in}

\maketitle
\thispagestyle{plain}
\pagestyle{plain}


\begin{abstract}
In the context of robotics, accurate ground-truth positioning is the cornerstone for the development of mapping and localization algorithms.
In outdoor environments and over long distances, total stations provide accurate and precise measurements, that are unaffected by the usual factors that deteriorate the accuracy of \ac{GNSS}.
While a single robotic total station can track the position of a target in three \ac{DOF}, three robotic total stations and three targets are necessary to yield the full six \ac{DOF} pose reference.
Since it is crucial to express the position of targets in a common coordinate frame, we present a novel extrinsic calibration method of multiple robotic total stations with field deployment in mind.
The proposed method does not require the manual collection of ground control points during the system setup, nor does it require tedious synchronous measurement on each robotic total station. 
Based on extensive experimental work, we compare our approach to the classical extrinsic calibration methods used in geomatics for surveying and demonstrate that our approach brings substantial time savings during the deployment.
Tested on more than \SI{30}{km} of trajectories, our new method increases the precision of the extrinsic calibration by \SI{25}{\%} compared to the best state-of-the-art method, which is the one taking manually static ground control points. 

\end{abstract}



\acresetall
\section{Introduction}

In mobile robotics, obtaining reference trajectories is vital for the development and evaluation of mapping and control algorithms~\cite{Sanchez-Cuevas2020}, while being critical to cornerstone datasets~\cite{Helmberger2022}.
In outdoor environments, total stations are the preferred choice to obtain high-accuracy measurements in the order of millimeters~\cite{Kalin2022}.
A total station is a precision measurement instrument equipped with optics that allow it to be precisely aimed at a given target.
Two angles (i.e., elevation and azimuth) and the range between the total station and the target are measured.
This information is sufficient to express the position of the target in the coordinate system of the total station.
They are not affected by the factors that may otherwise inhibit the usage of the \ac{RTK} \ac{GNSS} localization, such as dense foliage or urban and natural canyon environments~\cite{Kubelka2020}.
The only major requirement is the line-of-sight between a total station and the tracked target.
In the case of a static robotic platform with several targets attached to it, it is possible to obtain its six \ac{DOF} pose (i.e., position and orientation) as shown by \citet{Pomerleau2012}.
This procedure requires a series of manual measurements that capture each of these targets.
If the robotic platform moves during the measurement, a combination of a prismatic retro-reflector (i.e., simply \emph{prism} in the remaining of this article) and a \ac{RTS} is required~\cite{Cheng2011}.
The term \emph{robotic} in \ac{RTS} denotes the ability to automatically track a prism in motion.
Since a single \ac{RTS} can continuously track only one prism at a time, it is necessary to use at least three \acp{RTS} and three prisms to achieve the six \ac{DOF} pose tracking of a dynamic robotic platform. 
In recent years, manufacturers, such as Trimble, provide active prisms with infrared signature insuring that a \ac{RTS} will only track a given prism without being disturbed by other prisms in proximity.
This new feature allowed us to investigate novel solutions to reconstruct trajectories of moving vehicles.
\begin{figure}[t]
	\centering
	\includegraphics[width=0.9\linewidth]{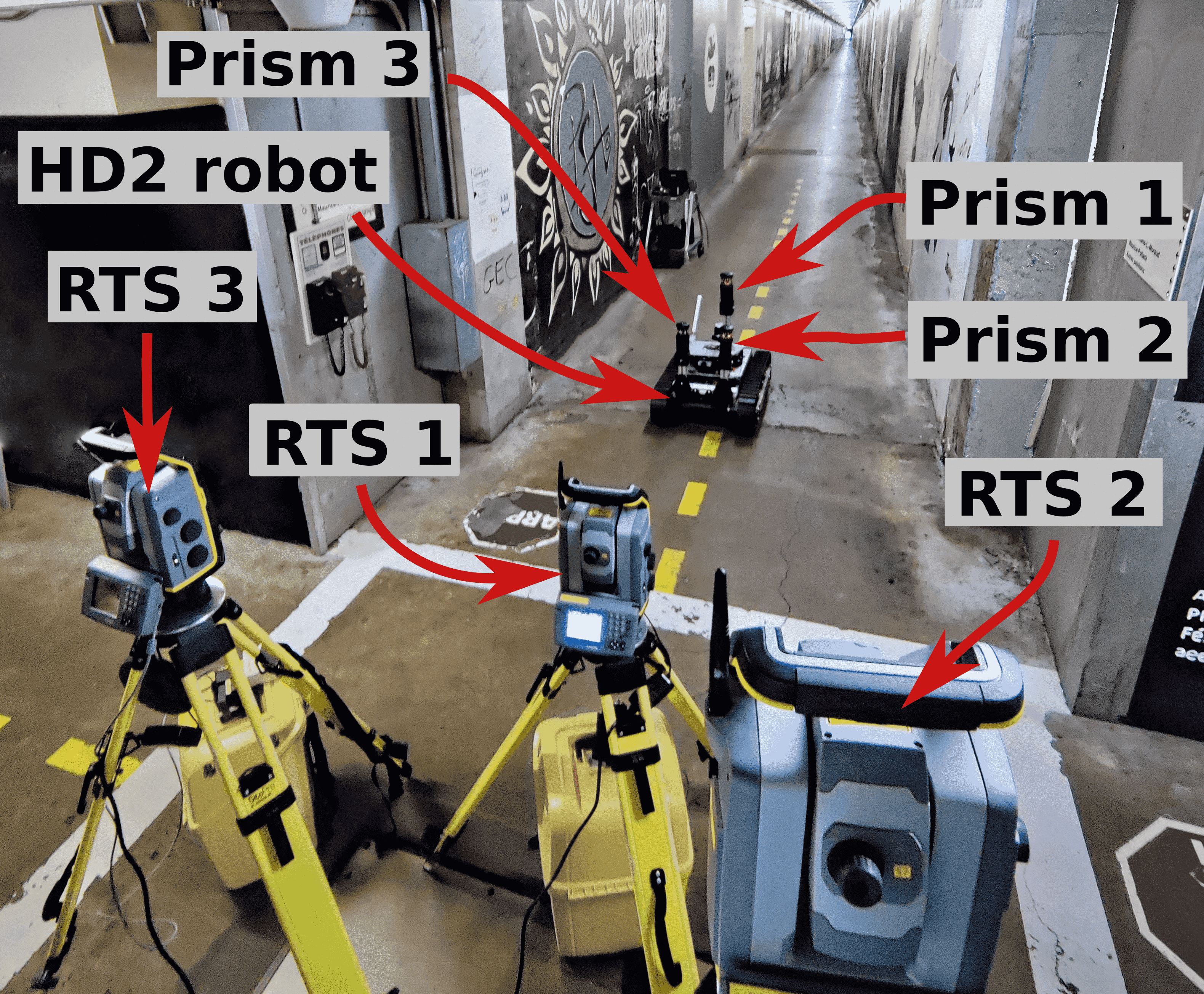}
	\caption{Setup to record reference trajectories using three robotic total stations to track three active prisms located on a HD2 robotic platform to reconstruct its six \acp{DOF} in a \SI{230}{m} tunnel deployment at Université Laval.}
	\label{fig:intro}
\end{figure}

As shown in \autoref{fig:intro}, each of the three \acp{RTS} is tracking its assigned prism that is mounted on the robotic platform and outputs the position of the associated prism expressed in its coordinate system.
To obtain the pose of a robot, there are two necessary conditions:
\textbf{1)}~All data must be expressed in a common coordinate frame based on an extrinsic calibration, also called \emph{resection} or \emph{free-stationing} in the surveying field.
In the literature, there are several extrinsic calibration methods for multiple \acp{RTS} based on static \acp{GCP}~\cite{Chukwuocha2018, Zhou2020}.
Yet, they require a significant effort in preparation and special equipment such as geodesic pillars to achieve the desired accuracy.
\textbf{2)}~The \ac{RTS} measurements need to be synchronized.
Contrary to position measurements from \ac{GNSS} receivers, the measurements are not executed synchronously between multiple \acp{RTS} by default since this functionality is usually not required in surveying~\cite{Thalmann2021}.
Therefore, temporal interpolation is necessary to exploit the data from these different \acp{RTS} to obtain the robotic platform's pose.

In this paper, we propose an extrinsic calibration method for three \acp{RTS} while a vehicle is in motion (i.e., without having to set up static reference points in the environment as done in our previous work~\cite{Vaidis2021}).
This method does not require manual registration of additional reference points, and the same configuration is used both for the calibration and the subsequent ground truth positioning, drastically reducing the setup time.
The method exploits the known distances between the prisms attached to the robotic platform and only requires the robot to be driven along a random trajectory throughout the experimental area.
We evaluate the proposed approach against existing extrinsic calibration methods for \ac{RTS} using a dataset consisting of more than \SI{30}{\km} of indoor and outdoor trajectories.
These datasets and the code are available online to the community.\footnote{\url{https://github.com/norlab-ulaval/RTS_Extrinsic_Calibration}}

\section{Related work}
\label{sec:related_work}

First, we describe extrinsic calibration methods found in the surveying literature for multiple-\ac{RTS} setups.
Then, we present works related to using multiple \ac{RTS} together. 
Finally, we list various robotic applications of \ac{RTS} used for acquiring reference trajectories of vehicles, and we put our work in this context. 

For all applications using multiple \ac{RTS}, measurements need to be expressed in a common coordinate frame.
The process of finding appropriate transformations is called extrinsic calibration.
The most common extrinsic calibration methods use multiple static \acp{GCP}.
The minimum number of required static \acp{GCP} is two, and this method is called \emph{two-point resection} in surveying~\cite{Milburn1987,Chukwuocha2018}.
This method requires the knowledge of the relative position of two \acp{GCP} with millimeter accuracy.
This requirement can be very difficult to comply with during field deployment.
Therefore, in most applications, three or more static \acp{GCP} with unknown global coordinates are used~\cite{Zhou2020}.
In outdoor environments with good \ac{GNSS} coverage, the \ac{GNSS} can be used to obtain the \ac{GCP} coordinates~\cite{Alizadeh2018}.
In that case, the extrinsic calibration expresses the pose of \ac{RTS} in the global frame of the \ac{GNSS}.
Although all of these methods with static \acp{GCP} are accurate in the order of a few millimeters, they can take hours to be carried out to achieve the desired accuracy~\cite{Merkle2004UseOT}.
To address this issue, a new extrinsic calibration method, which dynamically captured \acp{GCP}, was implemented by~\citet{Zhang2022}.
The \acp{GCP} were generated by two \acp{RTS} tracking one prism carried by an \ac{UAV}.
Although such measurements are less accurate than the static ones, the large number of \acp{GCP} obtained allows to compensate for the inaccuracy and provides a five-millimeter-accurate result in two minutes.
In this paper, a new dynamic extrinsic calibration that uses multiple prisms is presented, which does not need \acp{GCP}.

To properly analyze the results obtained by \acp{RTS}, it is necessary to take into account the different types of measurement noise. 
The first type of noise originates in extrinsic calibration.
The works of \citet{Horemuz2011, AminAlizadeh2018} searched for the optimal number of \acp{GCP} to minimize the uncertainty of the extrinsic calibration.
The method we propose removes the requirement of \acp{GCP} altogether while still providing a precise extrinsic calibration.
Another source of noise is the measurement equipment itself.
The contributing factors are the alignment of the prism with respect to the line of sight of the instrument and the type of electronic distance measurement unit inside the instrument.
Errors of two to four millimeters can occur~\cite{Lackner2016}.
Weather conditions also have a significant impact on range accuracy~\cite{TBAfeni2014}.
The differences in temperature and pressure need to be compensated as well~\cite{Rodriguez2021}.
In multiple-\ac{RTS} configurations, the temporal synchronization of the instrument clocks significantly affects the accuracy~\cite{Mao2018}.
An error of a one millisecond in the synchronization can lead to inaccuracy of one millimeter in the resulting measured position for a speed of \SI{1}{m.s^{-1}}. 
The usage of the \ac{GNSS}'s clock can mitigate this problem~\cite{Thalmann2021}.
Moreover, \ac{RTS} configurations that require communication over large distances can benefit from the long-range radio protocol with time synchronization~\cite{Vaidis2021}.
Finally, the last type of noise to be considered is interlinked with the application of tracking mobile robots.
The motion of the robotic platform can lead to outlier measurements.
Kalman filtering can be applied to the raw data to increase the precision as demonstrated by~\citet{Zhang2022}.
Some applications may require interpolation of the \ac{RTS} measurements, which adds another source of errors.
A simple linear interpolation can be used to process the data and synchronize them~\cite{Vaidis2021}. 
Although not used in surveying for interpolation, \acp{GP} are widely used in robotics to obtain continuous trajectories of robotic platforms and can be applied to prism trajectories~\cite{Anderson2015}.
In this paper, a new pre-processing pipeline applied to the raw \ac{RTS} data is introduced to increase the precision of the proposed extrinsic  calibration.

In mobile robotics, a wide variety of position-referencing systems are based on \acp{RTS}, but their use remains overall atypical.
An application of these systems is to register many robots in the same global frame before beginning swarm exploration of extreme environments~\cite{ebadi2022present}.
The design of these position-referencing systems also depends on the number of \ac{DOF} required, and also on the payload capacity of the platform carrying prisms.
Most of the referencing systems use only one \ac{RTS} to track the position of the robotic platform, being a skid steered robot~\cite{MacTavish2018}, a tracked robot~\cite{Kubelka2015}, a tethered wheeled robot~\cite{McGarey2018}, a planetary rover~\cite{Lemus2014}, an unmanned surface vessel~\cite{Hitz2015}, a \ac{UAV}~\cite{Schmuck2019} or a walking robot~\cite{Bjelonic2018}. 
Adding a second \ac{RTS} leads to a reduction in the uncertainty of the position as shown by~\citet{GabrielKerekes2018}.
In the work of~\citet{Reitbauer2020}, a compost turner with two different prisms attached to it was tracked by two \acp{RTS}.
This configuration provided ground truth measurement on four \ac{DOF}, namely the position and the yaw angle.
To obtain the full position and orientation reference of a robotic platform, it is possible to use a single \ac{RTS} with three targets attached to the platform~\cite{Pomerleau2012}.
The single \ac{RTS} was used to manually measure three different prisms while the platform remained static.
To the best of our knowledge, \citet{Vaidis2021} was the first proposed solution to track the full pose of a vehicle in a continuous manner.
To achieve this trajectory reconstruction, three \acp{RTS} coupled with three prisms, thus highlighting challenges related to extrinsic calibration required to transform all \acp{RTS} into a common reference frame.
In this article, we propose a new dynamic extrinsic calibration based on the inter-prism distances.
This new method can be applied in various types of environments and does not require any \acp{GCP}, thus saving time during the field deployments.

\section{Theory}
\label{sec:theory}

We first present the static calibrations used in surveying.
Then, we describe our new pre-processing pipeline applied on the raw \ac{RTS} data for dynamic tracking.
Finally, we introduce the new dynamic calibration we proposed.

\subsection{Static methods}
\label{sec:static_resection_methods}

\textbf{Two-point resection --}
The two-point resection~\cite{Milburn1987} is the first calibration method used as a standard referential solution.
The method requires two \acp{GCP} with precisely known relative position.
It exploits the geometry of a triangle formed by two \acp{GCP} and the \ac{RTS} to locate the \ac{RTS} in the coordinate frame defined by the \acp{GCP}.
The \ac{RTS} is assumed to be perfectly leveled, hence two points are enough to compute the solution which yields four \ac{DOF} that can be directly solved, the position and the yaw angle of the \ac{RTS}. 
To obtain accurate results, the relative position of the two known \acp{GCP} needs to be known in the order of millimeters or better.


\begin{figure}[htbp]
	\centering
	\includegraphics[width=0.8\linewidth]{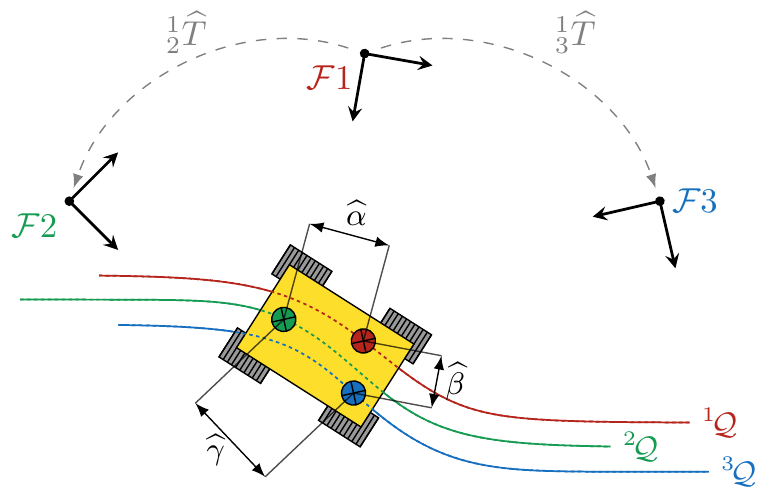}
	\caption{Notation used to define the cost functions minimized with the dynamic inter-prism calibration. The cost function minimizes the difference between the apparent inter-prism distance $\widehat{\alpha}$, $\widehat{\beta}$ and $\widehat{\gamma}$ and their real values. They are computed from the interpolated prism trajectories $^i\mathcal{Q}$ given by the raw data $^i\mathcal{P}$ being pre-processed by the pipeline. The result of the minimization are the rigid transformations $^1_2\widehat{\bm{T}}$ and $^1_3\widehat{\bm{T}}$ between the frame $\mathcal{F}_1$ of the first \ac{RTS} taken as the global frame, and the other two \ac{RTS} frames $\mathcal{F}_2$ and $\mathcal{F}_3$.}
	\label{fig:notation}
\end{figure}

\textbf{Static \acp{GCP} calibration --} This method requires at least three static \acp{GCP} ($n\geq3$).
A list of $n$ \ac{GCP} measurements ${}^i\mathcal{P}$ is collected by each \ac{RTS}, where $i\in\{1,2,3\}$ is the index of the \ac{RTS} used.
If the \ac{GCP} positions are also known in a global frame $\mathcal{F}_W$ (i.e., typically in \ac{GNSS} coordinates), a point-to-point alignment minimization is carried out between the global coordinates of the \ac{GCP} and their local coordinates in the \ac{RTS} frames, namely $\mathcal{F}_1$, $\mathcal{F}_2$ and $\mathcal{F}_3$  (see \autoref{fig:notation}).
The problem can be summarized as minimizing the point-to-point cost function as
\begin{align}
\label{eq:icp_p_to_point_min}
{}^W_i{\widehat{\bm{T}}} = 
\argmin_{\bm{T}} \: 
\sum_{k=1}^n (\bm{e}_{k}^{T} \bm{e}_k)
\text{,  with \hspace{0.01cm}}
\bm{e}_{k} = \bm{q}_k - \bm{T} \bm{p}_k\text{,}
\end{align}
where $\bm{q}_k$ is the $k$\textsuperscript{th} point of ${}^W\mathcal{P}$ representing the measured target positions taken as reference in the frame $\mathcal{F}_W$, $\bm{p}_k$ is the $k$\textsuperscript{th} point of ${}^i\mathcal{P}$ representing the positions of the targets in the \ac{RTS} frame $\mathcal{F}_i$, and $\bm{T} \in \SE(3)$ is a rigid transformation matrix used during the minimization.
The result given by \autoref{eq:icp_p_to_point_min} is a rigid transformation matrix ${}^W_i\widehat{\bm{T}} \in \SE(3)$ between the frame $\mathcal{F}_W$ and the \ac{RTS} frame $\mathcal{F}_i$.
Cartesian coordinates of each \ac{RTS} measurement can then be projected into $\mathcal{F}_W$ using ${}^W_1{\widehat{\bm{T}}}$, ${}^W_2{\widehat{\bm{T}}}$ and ${}^W_3{\widehat{\bm{T}}}$.
If the global coordinates of the \acp{GCP} are not known, it is possible to use the frame of the first \ac{RTS} $\mathcal{F}_1$ as the origin and simply solve for ${}^1_2{\widehat{\bm{T}}}$ and ${}^1_3{\widehat{\bm{T}}}$.

\subsection{Dynamic methods}
\label{sec:dynamic_methods}

\begin{figure*}[ht]
	\centering
	\includegraphics[width=0.90\linewidth]{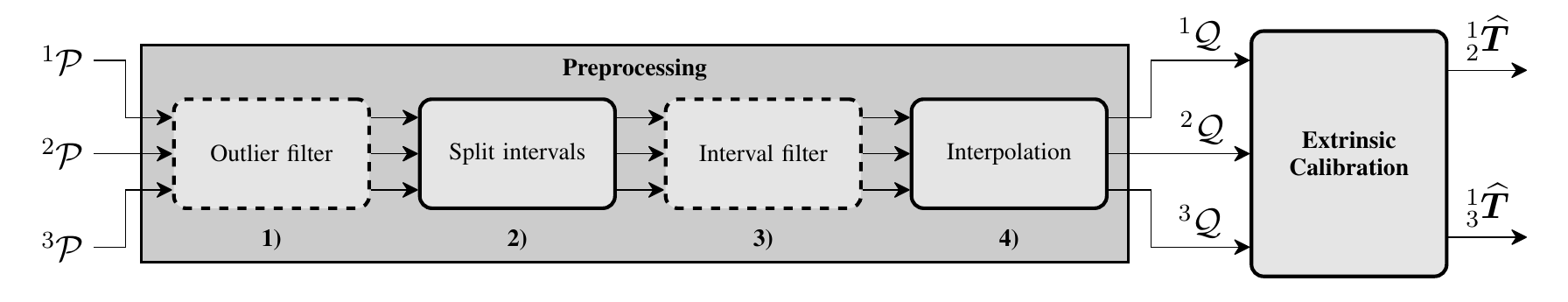}
	\caption{Pre-processing pipeline composed of four different modules applied on the raw \ac{RTS} data. The inputs ${}^i\mathcal{P}$ are lists of position measurements expressed in polar coordinates defined in the respective \ac{RTS} frames. Modules $2)$, and $4)$ (solid line) are necessary to process the \ac{RTS} data. The optional modules $1)$ and $3)$ (dotted line) are used to remove the outliers to increase the precision of the results. The output of the pre-processing pipeline are the interpolated prism trajectories ${}^i\mathcal{Q}$ expressed in Cartesian coordinates.
	They serve as the input to the extrinsic calibration block. 
	The calibration computes the rigid transformations ${}^1_2\widehat{\bm{T}}$ and ${}^1_3\widehat{\bm{T}}$ between the frame of the first \ac{RTS} and the frames of the second and third \ac{RTS}, respectively.}
	\label{fig:pre_process_pipeline}
\end{figure*}

\textbf{Pre-processing pipeline --} Methods for dynamic calibration use coordinates of moving prisms typically attached to a mobile robotic platform.
Therefore, it is beneficial to pre-process the \ac{RTS} measurements to limit the influence of outliers and to interpolate the data.
While the outlier removal improves the calibration accuracy, interpolation is required when the \acp{RTS} do not capture the target coordinates synchronously.  
The complete pre-processing pipeline we propose is shown in \autoref{fig:pre_process_pipeline}.
Inputs of the pipeline ${}^i\mathcal{P}$ are lists of raw polar coordinates (i.e., elevation, azimuth, and range) measured by the three \acp{RTS}.
The outputs $\mathcal{Q}$ are the filtered and interpolated Cartesian coordinates of the targets in their respective \ac{RTS} coordinate frame.
More precisely, the pipeline is composed of four distinct blocks:

\subsubsection{Outlier filter}

This module removes outliers that have too high derivative values of range, elevation, and azimuth when compared to their respective thresholds $\tau _r$, $\tau _e$ and $\tau _a$.
This filter also transforms the polar coordinates into Cartesian coordinates.

\subsubsection{Split intervals}

In the ideal case, each \ac{RTS} would provide an uninterrupted stream of prism position measurements.
However, due to obstacles or loss of a satisfactory lock onto the prism, the \ac{RTS} pauses the output until the lock is re-established with the required precision.
In such case, this module splits the trajectory into sub-intervals that exclude the outage.
It ensures that the time difference between two subsequent prisms positions is not greater than a threshold $\tau _s$.
Then, it keeps only the intersection of the sub-intervals found, \ie the time intervals when at least three \acp{RTS} were measuring uninterrupted.
The output is a list of sub-trajectories for each individual \ac{RTS}.

\subsubsection{Filter intervals}

Advance interpolation methods, such as \acp{GP}, can struggle when having only access to few supporting points.
Thus, this module is applied on intervals of valid sub-trajectories to mitigate this limitation.
The filtering criterion is their length; only intervals longer than a threshold $\tau _l$ are kept.

\subsubsection{Interpolation}

The interpolation is necessary for obtaining synchronized prism positions, as they are required by the Extrinsic Calibration module.
In this paper, we compare two types of interpolation: the linear interpolation and a \ac{GP}\footnote{Library \emph{Stheno}: \url{https://github.com/wesselb/stheno}} using an exponential quadratic kernel~\citep{Rasmussen2006}. 

The Split and Interpolation modules are necessary for our calibration method.
The rest of the modules (i.e., Outlier filer and Interval filter) are optional and increase the precision of the final results, as discussed in \autoref{sec:results}.


\textbf{Dynamic \acp{GCP} calibration --} This method is based on the same idea as the Static \acp{GCP}  calibration, with the difference of having \acp{GCP} collected dynamically, as proposed by \citet{Zhang2022}.
We assume \acp{RTS} are tracking dynamically the same prism, and their raw data ${}^i\mathcal{P}$ are pre-processed by the pipeline presented in \autoref{fig:pre_process_pipeline}.
This pipeline produces three lists of $n$ interpolated prism positions ${}^i\mathcal{Q}$ captured by each \ac{RTS} in its own frame $\mathcal{F}_i$.
We apply the same method as the static \acp{GCP} calibration with \ref{eq:icp_p_to_point_min}, with one of the three \ac{RTS} coordinates frames serving as the global frame.
The result is rigid transformations that express the data from the other two \acp{RTS} in the common global frame.


\textbf{Dynamic inter-prism calibration --} The final calibration method is based on the inter-prism distances constraint.
It allows three \acp{RTS} to track different prisms under the assumption that the relative positions between the prisms stay constant during the experiment.
This assumption is simple to implement as long as the robot is large enough for three prisms to be rigidly mounted onto the body.
Raw data from \acp{RTS} are also pre-processed following the pipeline of \autoref{fig:pre_process_pipeline}.
Based on the list of premeasured inter-prism distances $\Delta = \left\{\alpha,\beta,\gamma\right\}$ representing the distances between the prisms mounted on the robot, we can define an optimization task that finds the optimal rigid transformations between the \ac{RTS} frames (see \autoref{fig:notation}).

The optimization problem is defined as the minimization of the distance between the vector of apparent inter-prism distances $\widehat{\Delta} = \left\{\widehat{\alpha},\widehat{\beta}_,\widehat{\gamma}\right\}$ and their real values in $\Delta$.
The vector $\widehat{\Delta}$ is obtained from the trajectories $^1\mathcal{Q}$, $^2\mathcal{Q}$ and $^3\mathcal{Q}$, which depend on the rigid transformations in question. 
We define the parameters of the rigid transformations using the Lie algebra $\se(3)$ as $^W_i\bxi^\wedge$ \cite{Anderson2015}:
\vskip -1mm
\begin{equation}
    \begin{aligned}
    \label{eq:se3_variable}
    {}^W_i\bxi^\wedge =
		\begin{bmatrix}
			\bm{\rho} \\ \bphi
		\end{bmatrix}^\wedge
		=
		\begin{bmatrix}
			\bphi^\wedge & \bm{\rho} \\
			\vzero^T & 0
		\end{bmatrix}
		\, \text{with \hspace{0.05cm}}\, \bm{\rho}, \bphi \in \R^3, 
    \end{aligned}
\end{equation}
where $\bphi = [0,0,\phi], \phi \in \R$, as the \acp{RTS} are considered perfectly leveled.
The translations $\bm{\rho}$ and rotation $\bphi$ are a vector in $\se(3)$.
Following this definition, we can recompute ${}^W_i\bm{T}$ by taking the exponential map of $^W_i\bxi^\wedge$.
The cost function to minimize is then given by:
\begin{multline} 
    \label{eq:cost_function_2}
    ^1_2\widehat{\bxi}, {}^1_3\widehat{\bxi} = 
    \argmin_{{}^1_2\bxi, {}^1_3\bxi} 
    \frac{1}{3n} \sum_{j=1}^n
    \Biggl[
    \left(\norm{{}\bm{q}^j_1 - {}^1_2\bm{T}{}\bm{q}^j_2} - \alpha\right)^2 + 
    \\
    \left(\norm{\bm{q}^j_1 - {}^1_3\bm{T} \bm{q}^j_3} - \beta\right)^2 +
    \left(\norm{{}^1_2\bm{T}\bm{q}^j_2 - {}^1_3\bm{T}\bm{q}^j_3} - \gamma\right)^2
    \Biggr]
\end{multline}
where $\bm{q}_i^j$ is the $j^{\text{th}}$ point in ${}^i\mathcal{Q}$ and $\norm{\cdot}$ is the standard Euclidean norm. 
The minimization is performed by the iterative least squares method.
The resulting vectors $^1_2\widehat{\bxi}$ and $^1_3\widehat{\bxi}$ lead to the final rigid transformations $^1_2\widehat{\bm{T}}$ and $^1_3\widehat{\bm{T}}$.
As such, this method can get stuck in local minima, so we propose in the next paragraph an iterative method to identify a proper initial value to provide the optimizer.

\textbf{Search of prior --}
We developed a two-step iterative method based on the velocity of each points of ${}^i\mathcal{Q}$ to find a good prior for \autoref{eq:cost_function_2}.
The first step is to use a prior given by a point-to-point minimization to minimize \autoref{eq:cost_function_2} with the points considered static, which have less uncertainty, following a speed threshold defined as $\tau _v=\SI{1}{cm.s^{-1}}$.
Because the prism position are close enough to each other, their trajectories will be also close enough by applying this prior.
Subsequently, \autoref{eq:cost_function_2} is reused iteratively following an incremental step of $\SI{10}{cm.s^{-1}}$ for $\tau _v$ applied on the robot speed range to take more and more interpolated points for the minimization.
The rigid transformations used as priors are replaced by the successive ones obtained by the preceding $\tau_v$.
At the end of this first step, the inter-prism metric is used to find the best convergence obtained.
The second step is to redo the first step with the rigid transformations obtained by the best convergence of the first step as prior for all increment $\tau _v$.
At the end of this second step, the inter-prism metric is used again to find the best convergence obtained.
For both steps, an optimal convergence is validated if at least three other convergences have similar results (\ie having rigid transformations whose translation and angular differences are respectively less than \SI{5}{cm} and \SI{0.5}{deg} in our case).

\section{Experiments}
\label{sec:experiments}

The \acp{RTS} used in our experiments were three Trimble S7 tracking three Trimble MultiTrack Active Target MT1000 prisms.
The experimental setup was similar to the one presented by \citet{Vaidis2021} with an improved communication protocol.
The rate of the measurements was increased to \SI{2.5}{\hertz} per \ac{RTS} and the radio communication protocol was optimized for long-range measurements at \SI{1}{\km} distance.
In nominal conditions, the range measurement accuracy is \SI{2}{\mm} while the angular accuracy is \SI{1}{\arcsecond}.
To perform the experiments, a Clearpath Warthog mobile robot and a HD2 tracked platform from SuperDroid Robots were used in the outdoor and indoor experiments, respectively.
The three prisms were mounted on these robotic platforms (see \autoref{fig:intro} for the HD2 example).
Additionally, three \ac{RTK} \acp{GNSS} were installed on the Warthog mobile robot and used to gather data.
\begin{figure}[htbp]
	\centering
	\includegraphics[width=0.595\columnwidth]{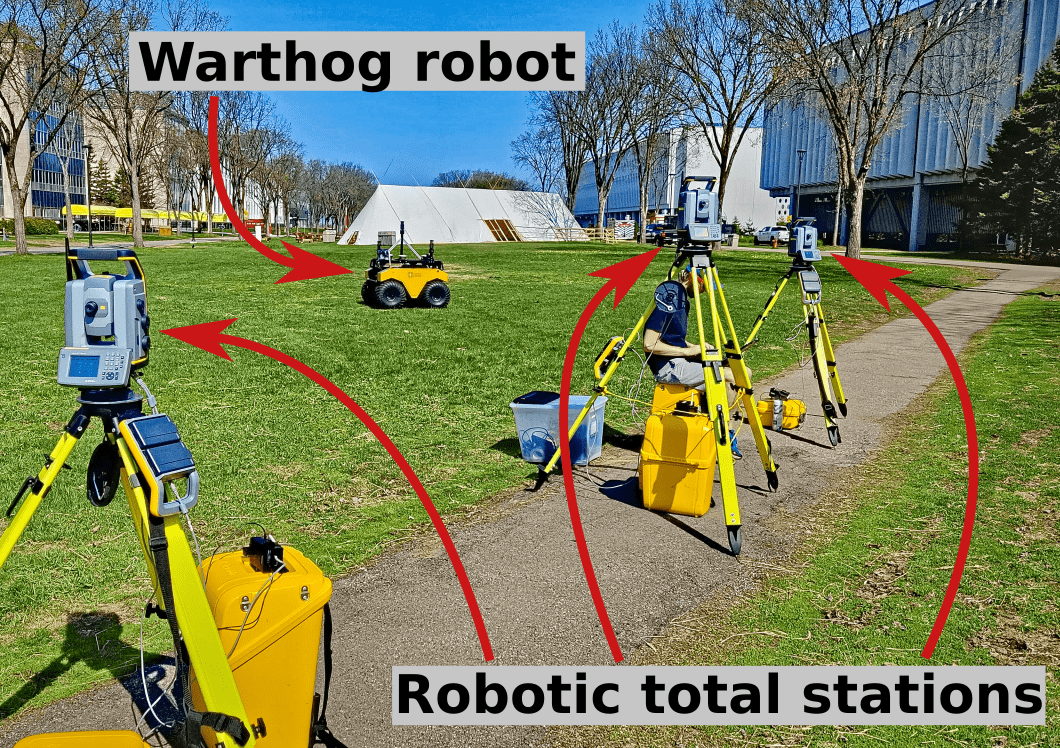}
	\includegraphics[width=0.385\columnwidth]{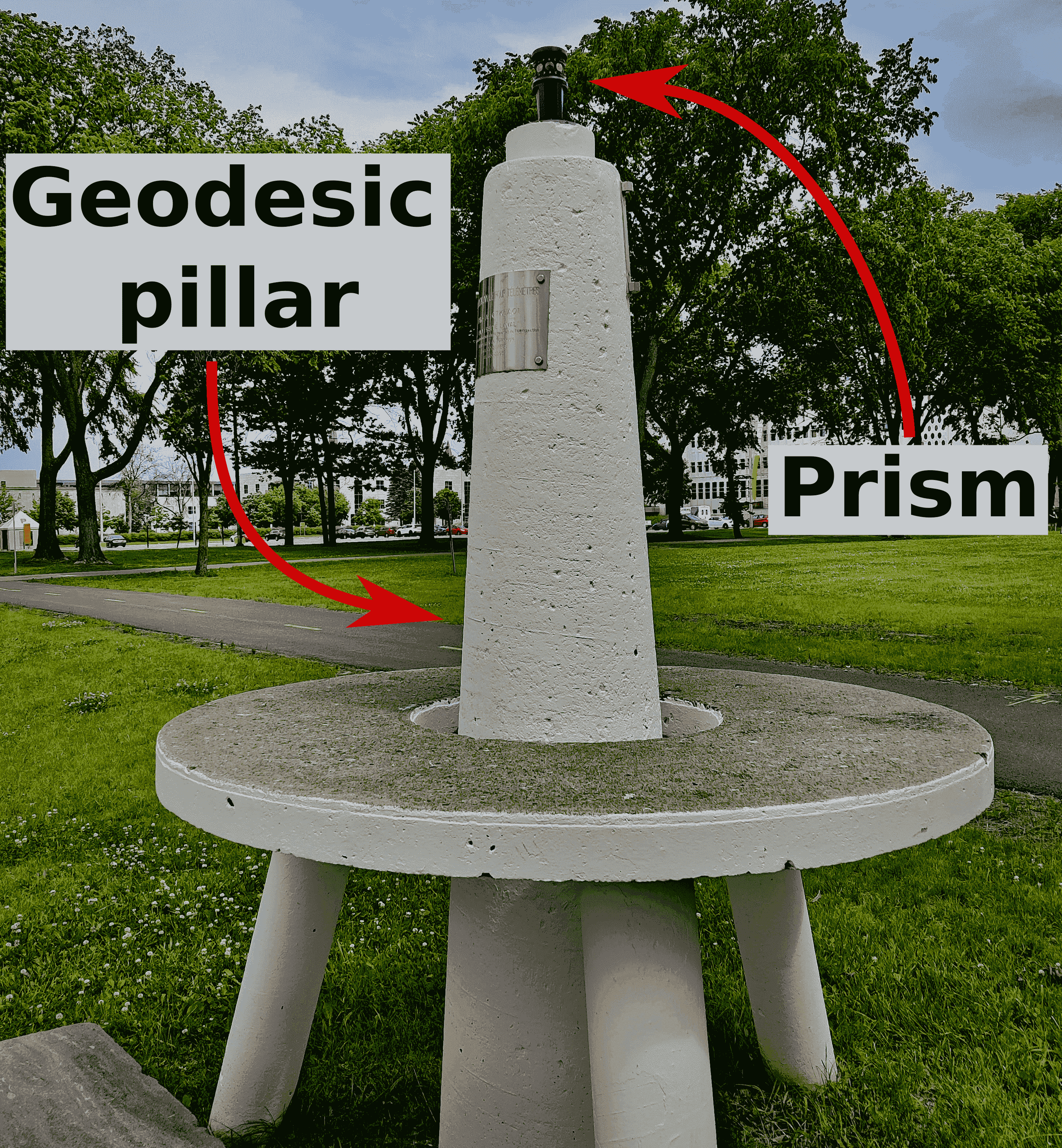}
	\caption{Large outdoor environment on the campus of Université Laval used for experiments. \emph{Left}: the setup of three \acp{RTS} tracking three prisms on
    the Warthog robotic platform. \emph{Right}: a prism mounted on one of four available geodesic pillars to perform the two-point resection method.}
	\label{fig:experiment_environments}
\end{figure}

In total, \SI{15}{} different deployments were carried out to evaluate the precision of the calibration methods presented in \autoref{sec:theory}.
The complete dataset consists of \SI{40}{} experiments, summing up to \SI{30.4}{\km} of prism trajectories.
The data were collected in two different types of environment at Université Laval, in tunnels and outdoors, as shown respectively in \autoref{fig:intro} and \autoref{fig:experiment_environments}.
Four geodesic pillars are located in the outdoor environment, which positions are known with a millimeter precision and surveyed every year by a specialized team. 
These pillars were used to evaluate the two-point resection method.
After each deployment, prism positions on the robotic platforms were measured by a single \ac{RTS} to compute the inter-prism distances $\Delta$.
For each prism position, ten measurements were averaged to reduce the impact of noise.

\section{Results}
\label{sec:results}

To evaluate our results, we use two different error measurements.
The first one is called the \acp{GCP} metric and is defined as the median distance between corresponding points after transforming them from their original coordinate frame to the global frame. 
Therefore, for $n$ \acs{GCP}, the \ac{GCP} metric will be the median of the $3n$ distances between the corresponding triplets of points measured by the \acp{RTS}.
The second metric called inter-prism metric is defined as the distance between $\Delta$ and $\widehat{\Delta}$.
The \acp{GCP} metric is considered to be the more accurate one because the measurements are static with less noise.
However, the inter-prism metric evaluate the precision of the results on the same set of data taken dynamically.
Therefore, the precision achieve by this metric is more pertinent to use for dynamic tracking of multiple prisms. 


\subsection{Sensitivity and ablation tests}
\label{sec:sensitivity_ablation}


To find the best parameter values for our pipeline presented in \autoref{sec:dynamic_methods}, we performed sensitivity tests.
We evaluated the impact of the thresholds $\tau_a$, $\tau_e$, $\tau_s$ and $\tau_l$.
The range threshold $\tau_r$ was excluded as it depends on the robot dynamics: $\tau_r$ was set to the maximum speed of the corresponding platform (\ie \SI{2}{m.s^{-1}} for the Warthog robot and \SI{1}{m.s^{-1}} for the HD2 robot).
The inter-prism metric was used to compare the results obtained with the different thresholds. 
These results were acquired using the static \ac{GCP} calibration to avoid the bias that would be introduced with the dynamic calibration methods which use the processed data.
Both the tunnels and the outdoor environment were used for this analysis.
The results have shown that the values of $\tau_a$, $\tau_e$ and $\tau_l$ have little impact on the results.
Indeed, our \acp{RTS} only produced a few outliers caused by errors in the measured angles.
Similarly, there were not too many \ac{RTS} outages in the data, thus majority of the sub-intervals are sufficiently long.
Therefore, we set the values to $\tau_a = \SI{1}{deg.s^{-1}}$ , $\tau_e = \SI{1}{deg.s^{-1}}$  and $\tau_l = \SI{6}{s}$ following reasonable physical quantities given our equipment.
On the other hand, the sensitivity test for $\tau_s$ give us a minimum error which is reached for a value of $\tau_s = \SI{1}{s}$.
In our experimental setup, this translates to cutting the trajectory if there are more than two missing data points in a row.

\begin{figure}[htbp]
	\centering
	\includegraphics[width=\columnwidth]{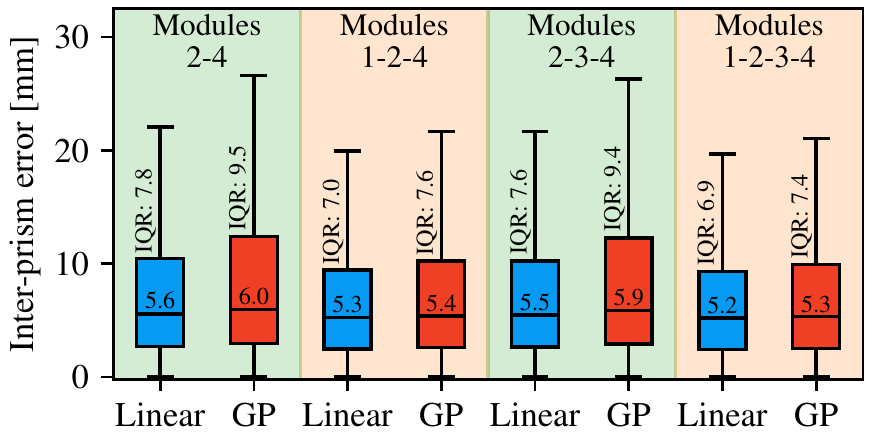}
	\caption{Error resulting from the ablation tests of different modules presented in \autoref{sec:dynamic_methods}. 
	The linear interpolation denoted by blue and the \ac{GP} by red.
	Median error are written in the middle of the box and the \acf{IQR} on the whiskers.}
	\label{fig:ablation_impact_pipeline}
\end{figure}

After determining the pipeline parameter values, we performed ablation tests over the pipeline modules to study their impact on the calibration accuracy.
The tests used the same experimental data and the same metric as the parameter search.
We compared the minimum necessary set of the modules $2)$ and $4)$, to the results achieved by adding the outlier filter module $1)$ and the interval filter module $3)$.
In parallel, we compared the performance of the linear interpolation against the \ac{GP} option.
The results are shown in \autoref{fig:ablation_impact_pipeline}.
First, we observe that \ac{GP} shows worse performance than the linear interpolation.
The median error is equivalent, but the \ac{IQR} is approximately one to two millimeters larger.
The probable cause is the limited amount of training data available in the sub-intervals that prevent the \ac{GP} from precisely interpolating the prism position.
The ablation test shows that the outlier filtering module positively affects the results by decreasing the median error by \SI{7}{\%}, and the error \ac{IQR} by \SI{15}{\%}.
It also confirms the parameter sensitivity analysis conclusions: in our dataset, the interval filtering module does not have a high impact.
It decreases the median error and the \ac{IQR} by approximately \SI{2}{\%}.
Finally, the complete pipeline decreases the median error by \SI{9}{\%}, and the error \ac{IQR} by \SI{18}{\%}, compared to the unfiltered \ac{RTS} data.
Additionally, we further consider only the linear interpolation in the calibration comparison.

\subsection{Calibration comparison}
\label{sec:comparison}

\begin{figure}[htbp]
	\centering
	\includegraphics[width=\columnwidth]{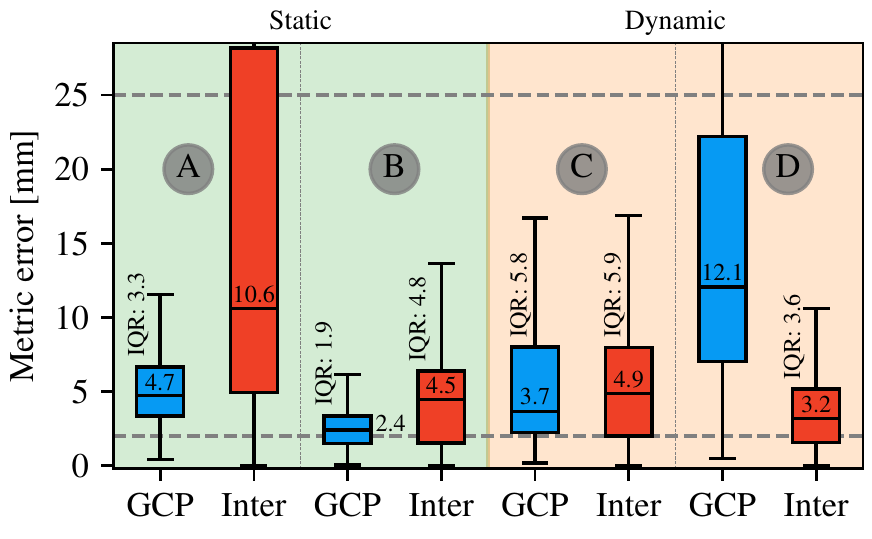}
	\caption{Comparison of the calibrations methods. Both the median error and the error \ac{IQR} are shown for the \acp{GCP} metric (blue) and the inter-prism metric (red). \emph{A}: Two-point resection. \emph{B}: Static \acp{GCP} calibration. \emph{C}: Dynamic \acp{GCP} calibration. \emph{D}: Dynamic inter-prism calibration. The upper dashed line represent the \ac{RTK} \ac{GNSS} precision and the lower dashed line represent the \ac{RTS} noise level.}
	\label{fig:comparison_resection_method}
\end{figure}

The results of the comparison between all presented calibration methods are shown in \autoref{fig:comparison_resection_method}.
Both the \acp{GCP} and the inter-prism metric were used for the comparison.
The two-point resection \emph{(A)} is less accurate than the static \ac{GCP} calibration \emph{B} according to both metrics.
Moreover, the dynamic \ac{GCP} calibration \emph{(C)} is better than \emph{(A)}, but more noisy than the static \ac{GCP} calibration \emph{(B)}, as also noticed by \citet{Zhang2022}.
On the other hand, the dynamic inter-prism calibration \emph{(D)} result strongly depends on the metric type.
Compared to the calibration \emph{(B)}, the \acp{GCP} metric yields a high median error of \SI{12.1}{mm}.
However, the inter-prism metric indicates that the dynamic inter-prism calibration \emph{D} also decreases the median and \ac{IQR} error by \SI{29}{\%} and \SI{25}{\%} respectively, compared to \emph{B}.
The best result in the inter-prism metric from \emph{(D)} is explained by the fact that this metric is directly minimized by \autoref{eq:cost_function_2}.
On the other hand, a difference of \SI{7}{mm} in translation and \SI{0.01}{deg} for the yaw angle was observed between the results of \emph{(B)} and \emph{(D)}.
Combined, these values give the error difference for the \acp{GCP} metric.
Since the noise level of the measurements is of the order of \SI{2}{mm}, these convergence differences could be attributed to the minimization. 
%
Compared to the inter-\ac{RTK} \ac{GNSS} distance measured during the outdoor experiments with the Warthog, the results coming from the calibration methods have a better precision as seen in \autoref{fig:comparison_resection_method}.

\section{Conclusion}
\label{sec:conclusion}

We have proposed a new dynamic extrinsic calibration method and a pipeline for pre-processing the \acp{RTS} data.
This new dynamic calibration exploits the inter-prism distance measured by a setup of three \acp{RTS} during a robotic deployment to compute the rigid transformations between the frames of different \ac{RTS}.
Moreover, our new calibration method does not rely on \acp{GCP} to be measured on the field, which saves from 20 to 45 minutes at the beginning of each robotic deployment.
Even with a higher error compared to other tested methods, the Dynamic inter-prism calibration still produce better results than a \ac{RTK} solution in optimal condition (i.e., open sky).
Additionally, the new pre-processing pipeline increased by \SI{18}{\%} the precision of the results, alongside the inter-prism calibration, which increased it by \SI{25}{\%} compared to the best state-of-the-art method. 
As field experiment can be messy, if \acp{GCP} were forgotten or mishandled, the Dynamic inter-prism calibration can also be used to recover the calibration of the \acp{RTS} as it does not require a different set up than the one used to reconstruct a six \ac{DOF} trajectory.

Throughout the experiments, a limiting condition to our calibration method has been found to be long straight lines or L-shaped trajectories. 
The lack of rotation in these trajectories under-constrained the minimization. Thus causing the results to be off by more than \SI{5}{cm} for the \ac{GCP} metric for the HD2's tunnel experiments even though the millimeter order was achieved for the inter-prism metric.
Based on our experience, another possible contributor to unsatisfactory results is the distance between the prisms on the HD2, which was around \SI{35}{cm}, compared to the one on the Warthog which respect the minimum distance of \SI{80}{cm} advised by Trimble.
For future work, we hypothesize that these short distances have had an impact on the experiment in the tunnel and must be further investigated.



\section*{Acknowledgment}

We thank Christian Larouche, director of the metrology laboratory at Université Laval, for his advises and providing us the precise coordinates of the geodesic pillars.
This research was supported by the  Natural  Sciences and Engineering  Research  Council of  Canada  (NSERC)  through the grant CRDPJ 527642-18 SNOW (Self-driving Navigation Optimized for Winter).


\IEEEtriggeratref{6}
\IEEEtriggercmd{\enlargethispage{-0.1in}}

\printbibliography

\end{document}